\documentclass{article}
\usepackage{graphicx} 
\usepackage{amsmath}
\usepackage{amssymb}

\title{Deep Reinforcement Learning Agents for Strategic Production Policies in Microeconomic Market Simulations}
\author{Eduardo C. Garrido-Merchán, Maria Coronado-Vaca, \\Álvaro López-López, Carlos Martinez de Ibarreta \\ Universidad Pontificia Comillas, Madrid, Spain \\ \texttt{ecgarrido@comillas.edu, mcoronado@icade.edu} \\ \texttt{allopez@comillas.edu, charlie@comillas.edu}}

\date{November 2024}

\begin{document}

\maketitle

\begin{abstract}
Traditional economic models often rely on fixed assumptions about market dynamics, limiting their ability to capture the complexities and stochastic nature of real-world scenarios. However, reality is more complex and includes noise, making traditional models assumptions not met in the market. In this paper, we explore the application of deep reinforcement learning (DRL) to obtain optimal production strategies in microeconomic market environments to overcome the limitations of traditional models. Concretely, we propose a DRL-based approach to obtain an effective policy in competitive markets with multiple producers, each optimizing their production decisions in response to fluctuating demand, supply, prices, subsidies, fixed costs, total production curve, elasticities and other effects contaminated by noise. Our framework enables agents to learn adaptive production policies to several simulations that consistently outperform static and random strategies. As the deep neural networks used by the agents are universal approximators of functions, DRL algorithms can represent in the network complex patterns of data learnt by trial and error that explain the market. Through extensive simulations, we demonstrate how DRL can capture the intricate interplay between production costs, market prices, and competitor behavior, providing insights into optimal decision-making in dynamic economic settings. The results show that agents trained with DRL can strategically adjust production levels to maximize long-term profitability, even in the face of volatile market conditions. We believe that the study bridges the gap between theoretical economic modeling and practical market simulation, illustrating the potential of DRL to revolutionize decision-making in market strategies.
\end{abstract}

\section{Introduction}
The complexity of developing effective business strategies and production policies in dynamic markets has long been recognized as a significant challenge, as highlighted by studies on business strategy and simulation in complex environments \cite{campbell2002business, jahangirian2010simulation}. In particular, in highly volatile and complex markets dealing with lots of explanatory variables of an endogenous variable, conventional assumptions of stability and predictability are often invalid, leading to chaotic conditions where traditional economic models struggle to provide accurate forecasts and actionable insights \cite{trippi1994chaos, pindyck2018microeconomics}. Given this complexity in the patterns involving variables such as how many production should a company perform in a certain time, it must be emphasized that neural networks, known for their capability as universal approximators, offer a potential solution by modeling non-linear patterns within the markets, enabling us to represent in the weights of the network intricate dependencies that are typically not representable by the capacity of simpler models like generalized linear models \cite{lecun2015deep}. However, neural networks trained by supervised learning need an annotated dataset, independent and identically distributed data  and a derivable numerical continuous loss function, assumptions that are often not possible to satisfy in daily markets.  

On the other hand, reinforcement learning, particularly in a simulated environment, provides a framework for agents to learn optimal strategies through experience, iteratively refining their actions based on feedback \cite{sutton2018reinforcement} that can be given at any moment in the training of the agent and can be discrete. In particular, this process mimics real-world learning by trial and error, where agents adjust their policies to maximize cumulative rewards over time. However, in scenarios where the simulator does not perfectly replicate real-world conditions like the markets given their complexity, introducing controlled noise, a process which is known as domain randomization, can help bridge this reality gap. By perturbing certain parameters within the simulator, it is possible to generate variations that cover a wide range of possible real-world scenarios. Within these perturbed distributions, one instance may closely approximate actual market conditions, improving the robustness of the agent’s learned policy \cite{tobin2017domain,guitta2023learning}.

In our approach, we combine the virtues of the representative capacity of deep learning networks and the way that an agent learns with reinforcement learning to leverage deep reinforcement learning (DRL) within a corrupted simulation environment to approximate optimal production policies of an agent in a market. In particular, DRL’s ability to handle high-dimensional spaces and learn complex policies enables us to model nuanced production strategies that adapt to fluctuating market dynamics, offering potential improvements in decision-making under uncertainty \cite{franccois2018introduction}. By iterating through multiple instances of a noisy simulator, our model develops robust strategies that can potentially be used in real, unpredictable market conditions, aiming to outperform static or traditional policy models.

The remainder of this paper is organized in the following way: First, Section 2 provides a review of the state-of-the-art in Deep Reinforcement Learning (DRL), including general approaches, applications in finance and economics, and techniques like domain randomization. Then, Section 3 illustrates the methodology, detailing foundational aspects of reinforcement learning and specific techniques in DRL. Afterwards, Section 4 describes the market design, including the simulation environment and the Markov Decision Process (MDP) formulation used to model agent interactions. Once that everything is defined, in Section 5, we present experimental results and evaluate the impact of producer actions on the simulated market. Finally, Section 6 concludes the paper and discusses potential avenues for future research.
\section{State-of-the-art}
In recent years, DRL has emerged as a powerful tool for solving complex decision-making problems in a plethora of domains. Concretely, DRL approaches have been successful in areas requiring sequential decision-making and adaptability such as the one described in this work, driven by the ability of neural networks to approximate complex functions and optimize actions through continuous feedback. For a complete description of DRL methodologies, far beyond the reach of this article, there is information in books such as those by \cite{sewak2019deep} and \cite{dong2020deep}, that provide comprehensive overviews of DRL techniques and their wide range of applications.

Dealing with the application of this article, microeconomics, and with the fact that no papers explore the production of items,  it is interesting to see a related application of DRL where it is widely used, finance, especially for tasks such as portfolio management, asset pricing, and trading strategies. Several surveys, including \cite{fischer2018reinforcement}, \cite{sahu2023overview}, and \cite{hambly2023recent}, have reviewed the current applications of DRL in finance, denoting how DRL can handle the stochastic nature of financial markets and adapt strategies to maximize returns, which is something similar to what we are going to illustrate in this paper. These studies emphasize the potential of DRL in transforming financial decision-making, where agents can learn optimal trading strategies by simulating market environments and optimizing based on performance feedback. However, they also highlight the challenges in applying DRL to finance, such as handling noisy data, high-dimensional spaces, and maintaining robustness in volatile markets. In particular, a specific interesting subfield where DRL is being applied is environmental, social, and governance (ESG) finance, which focuses on sustainable and socially responsible investments. Recent work by \cite{garrido2023deep} explores the integration of DRL with ESG criteria, exploring how DRL models can prioritize investment strategies that align with sustainability goals.

Dealing with economics, many current DRL models in economics address tangent theoretical or predictive problems. According to \cite{mosavi2020comprehensive}, the integration of DRL in economic applications is primarily constrained by the complexity of modeling economic actions that involve multiple, interdependent factors, and by the challenge of capturing realistic producer behaviors in simulation environments.
Although several DRL approaches are beginning to be used in economic problems no one deals with the actions that can be done by a producer \cite{mosavi2020comprehensive}. Consequently, we target this research gap in this article with our proposed approach whose methodology is going to be explained in the following section.

\section{Methodology}
In this section we briefly describe the fundamentals of, first, reinforcement learning and, then, of deep reinforcement learning to illustrate why the use of this methodology is adequate to estimate a policy of a production agent in a market simulation.
\subsection{Reinforcement Learning}
Reinforcement learning \cite{sutton2018reinforcement} is a learning paradigm in which an agent interacts with an environment in discrete time steps, with the purpose of learning a policy that maximizes a given loss function, like cumulative rewards or, in our case, cumulative profits in a time period. Formally, the interaction of the agent with the environment is modeled as a Markov Decision Process (MDP), defined by a tuple \((\mathcal{S}, \mathcal{A}, P, R, \gamma)\), where \(\mathcal{S}\) is the set of observed states by the agent, \(\mathcal{A}\) is the set of actions that the agent can perform, \(P(s'|s, a)\) is the state transition probability, representing the probability of moving to state \(s'\) from state \(s\) after taking action \(a\), \(R(s, a)\) is the reward function, representing the immediate reward after taking action \(a\) in state \(s\) that the agent perceives and \(\gamma \in [0, 1]\) is a discount factor variable, determining the importance of future rewards. The goal in RL is to find an optimal policy \(\pi^*\) that maximizes the expected cumulative reward of the agent in its trajectory through states, or the return \(G_t = \sum_{k=0}^{\infty} \gamma^k R(s_{t+k}, a_{t+k})\), where $t$ is the timestep. This optimal policy is typically obtained by maximizing the value function \(V^{\pi}(s)\), defined as:
\begin{equation}
V^{\pi}(s) = \mathbb{E}_{\pi} \left[ G_t | s_t = s \right]
\end{equation}
and the action-value function \(Q^{\pi}(s, a)\):
\begin{equation}
Q^{\pi}(s, a) = \mathbb{E}_{\pi} \left[ G_t | s_t = s, a_t = a \right].
\end{equation}
The agent seeks to maximize \(V^{\pi}(s)\) and \(Q^{\pi}(s, a)\) through various RL algorithms. However, reinforcement learning stores the value function and action-value function in a table that maps actions with states. In particular, if the observation space or action space of the MDP of an agent is high-dimensional or continuous, the table will need a huge amount of memory to store all the values mapping states and actions by the reward and computational resources to be traversed by the agent in training time. In order to circumvent the issue of the memory and assuming a smoothness between similar states and actions, Deep Reinforcement Learning will use deep neural networks to represent this mapping, as we will see in the following section. 

\subsection{Deep Reinforcement Learning}
As we have introduced in the previous subsection, DRL generalizes traditional RL by using neural networks as function approximators, particularly useful in high-dimensional spaces where traditional tabular methods are infeasible \cite{franccois2018introduction}. In DRL, the agent employs a deep neural network \(Q(s, a; \theta)\) with parameters \(\theta\) to approximate the Q-values or directly model the policy \(\pi(a|s; \theta)\). Roughly, depending on whether they approximate the value function, they directly optimize the policy or if they combine both approaches we find a family of methods belonging to where they focus the learning that we summarize here. In this work, we will use three algorithms, one belonging to each different type to test how they adapt to this problem. 

The methodology for deep reinforcement learning (DRL) encompasses several approaches, including value-based, policy-based, and actor-critic methods, each with unique strategies for optimizing agent behavior within a complex environment.

In a value-based approach, a popular algorithm is Deep Q-Learning, known as DQN \cite{mnih2013playing}. DQN uses Q-learning combined with neural networks to approximate the action-value function \( Q(s, a) \), which estimates the expected reward for taking an action \( a \) in a state \( s \). The objective is to learn an optimal Q-function, \( Q^*(s, a) \), that satisfies the Bellman equation:
\begin{equation}
Q^*(s, a) = \mathbb{E} \left[ R(s, a) + \gamma \max_{a'} Q^*(s', a') \mid s, a \right],
\end{equation}
where \( R(s, a) \) is the immediate reward, \( \gamma \) is the discount factor, and \( s' \) is the next state. DQN updates its parameters by minimizing a loss function $L(\theta)$ like the mean squared error (MSE) between a predicted Q-value and a target Q-value. Finally, the policy is set as the actions that are selected based on a greedy approach, that is selecting the action with the highest probability associated, or an \( \epsilon \)-greedy policy, relaxing the selection with a probability $\epsilon$ of choosing any action, to balance exploitation of learned actions and exploration of new possibilities.

Another family of algorithms, Policy-based methods, such as Proximal Policy Optimization (PPO) \cite{schulman2017proximal}, focus on directly optimizing the policy \( \pi(a | s; \theta) \) to maximize the expected cumulative reward. PPO uses a clipping mechanism to maintain stable updates, keeping the policy within a trust region to avoid unstable learning. Concretely, the PPO objective function is defined as:
\begin{equation}
L^{\text{PPO}}(\theta) = \mathbb{E}_t \left[ \min \left( r_t(\theta) \hat{A}_t, \text{clip}(r_t(\theta), 1 - \epsilon, 1 + \epsilon) \hat{A}_t \right) \right],
\end{equation}
where \( r_t(\theta) = \frac{\pi(a_t \mid s_t; \theta)}{\pi(a_t \mid s_t; \theta_{\text{old}})} \) is the probability ratio between the new and old policies to avoid unstable learning, \( \epsilon \) is a small constant, and \( \hat{A}_t \) is the advantage function, which indicates the relative value of an action in a state with respect to the value function of the state. Critically, the clipping mechanism in PPO ensures that updates remain conservative, preventing large jumps in policy space that could destabilize learning.

Finally, both policy-based and value-based methods can be combined, something that is known as actor-critic methods. For example, the A2C method, Advantage Actor-Critic \cite{mnih2016asynchronous},  maintains two models: an actor which is policy-based and directly learns the policy. On the other hand a critic (value-based) estimates the value function, that is, the expected future reward of each action. Combining both neural networks, the actor is responsible for deciding the actions based on the feedback provided by the critic modeling an advantage function, which accelerates and stabilizes the training and balances exploration and exploitation.

Having briefly covered the fundamental concepts of DRL, we now provide details in the following section of the market design of our experiment where the agent is trained using the DRL algorithms that we have shown in this section.

\section{Market Design}
In this section, we will provide all the details of the market simulation that we have coded in a Gym environment. We will first describe the dynamics of the market and the parameters of the environment and then provide information about the observation space, action space and reward of the producer agent. The code of all the simulation and the experiments is available at Github (\texttt{https://github.com/EduardoGarrido90/micro\_agents}).

\subsection{Market Simulation}
We first describe the Market simulation where we have trained several agents. It basically contains code that simulates the dynamics of a market involving one particular asset. We assume that this item is unique in the market, in the sense that no substitute or complementary products exist. However, we simulate the behaviour of competitors producing the same item as the agent's producer.

The simulator extends the environment class of the gymnasium library, which includes a constructor, reset, step and render method. In the constructor, we initialize the production limit per producer, the minimum production, the number of competitors and their initial produced quantities sample from a discrete uniform distribution in the ranges of production, we also initialize the discrete action space of our agent, which are the units produced and the observation space of our agent, that is a Box space involving the variables detailed in the following subsection. We also initialize the market price of the asset, total supply and demand, fixed costs of the company sampled from a normal distribution, the coefficients of the cubic total production curve, the timestep of the simulation equal to 0 and some variables that the agent will use to track the production of the competitors in previous timesteps. The value of the parameters that we have used as a sample to configure the environment are the ones illustrated in Table \ref{table:market_simulator_parameters}.

\begin{table}[h!]
\centering
\begin{tabular}{|l|l|p{9cm}|}
\hline
\textbf{Parameter Name} & \textbf{Value} & \textbf{Description} \\ \hline
Minimum production & 0 & Minimum production per producer in the simulation \\ \hline
Production limit per producer & 14 & Maximum production limit per producer \\ \hline
\# competitors & 3 & Number of competing producers in the market \\ \hline
Initial price & 15.0 & Starting price of goods in the market \\ \hline
Max fixed costs & 10.0 & Maximum fixed production costs per producer \\ \hline
Min fixed costs & 1.0 & Minimum fixed production costs per producer \\ \hline
Total prod. constant coef. & 0.0 & Coefficients of the cubic production curve \\ \hline
Total prod. linear coef. & 4.0 & Coefficients of the cubic production curve \\ \hline
Total prod. quadratic coef. & -0.6 & Coefficients of the cubic production curve \\ \hline
Total prod. cubic coef. & 0.03 & Coefficients of the cubic production curve \\ \hline
Elasticity & 1.02 & Elasticity factor affecting demand responsiveness \\ \hline
Base demand & 43.4 & Base demand calculated as production limit $\times$ 3.1  \\ \hline
Production noise & 0.05 & Noise factor added to production values \\ \hline
Storage factor & 2 & Factor affecting storage penalty due to excess production  \\ \hline
Max brand effect & 0.3 & Maximum profit increase (30\%) from brand influence  \\ \hline
Max subsidy & 10.0 & Maximum subsidy level at high production levels \\ \hline
\end{tabular}
\caption{Parameter values for a sample of the Market Simulator Experiment}
\label{table:market_simulator_parameters}
\end{table}

However, we emphasize that these values are only initial point estimates and that they are all a sample of different random variable whose joint random variable is the distribution of valid market environments where the DRL agents can be trained successfully to obtain competitive policies with respect to baseline strategies.

Once the simulator has been initialized, the step method emulates a timestep in the simulation, obtaining the action performed by the agent as the result of its previous observed state and giving as a result the new state of the market, the reward obtained by the agent which is going to be the profit and several traces for log purposes.   

The step method first simulates a random walk of the units produced by the competitors as a result of the action of the agent, if the agent produces more, the supply grows and the price will go down for the same fixed costs, so the competitors try to produce less and the contrary happens if the agent produces less. All quantities are clipped from zero to the maximum number of units that can be produced and the total supply of the item in the market is computed as the sum of the units produced by the agent and the competitors. Then a base demand is simulated as 3 times the production limit per the producer, summed by Gaussian random noise. This base demand is substracted a quantity which is based in the price of an item, more expensive less demand and a elasticity, representing that if the price rises a lot the demand is going to decrease quadratically up to certain elasticity. We also simulate a sinusoidal variation in the demand every 200 steps approximately representing stationary changes in the demand of the asset that the agent may learn having the timestep as one of the variables observed.

The total production cost curve is modelled as a cubic equation whose beta coefficients are set in the constructor as a function of the ranges of price of the asset. The cubic equation models a decrease in the price as a result of economies of scale when we produce more units of the item and finally a dramatic rise in the price if lots of items are produced as raw materials would be scarce and imported from abroad. The cost is also perturbed by random noise, modeling operational risk events as in fixed costs and other variables of the simulator. The supply variation of the company of the agent is also sinusoidal and changes the production cost. Moreover, if the producer quantity surpasses the total demand, storage factor penalizations are sum to the production cost exponentially. 

The price of the asset changes with respect to a demand and supply curve, if more units are produced than demanded the price of the asset becomes lower and the contrary thing happens if demand is lower than supply. 

The revenue, which is the reward of the agent, is computed as the price of the items times the quantity produced, the income, substracted by the fixed costs of the company, perturbed by random noise representing operational risk and the production cost. Moreover, a brand effect as a result of market cap is introduced, so producing more items increases the revenue as a result of image brand, also perturbed by random noise representing the volatility of the market. Finally, a subsidy of the public administration is given to companies based on a production level. The final profit is set as the reward of the agent and the observation includes all the variables that in reality could be seen by a company, as it is detailed in the following subsection.

\subsection{Markov Decision Process Design}
In this section we describe the variables that the agent deals with in order to estimate a production policy in this market as the ones that are contained in a Markov Decision Process \cite{puterman1990markov}, that is, the variables belonging to the observation space, the action space and the reward.

We have summarized the variables perceived by the agent in the mentioned simulation in Table \ref{table:agent_spaces}. Concretely, the table presents the components of the agent's observation space, action space, and reward function of the described market simulation in the previous subsection. Each element is categorized by its name, role, variable type, and description, detailing how it contributes to the agent's policy. In particular, the observation space includes variables that allow the agent to perceive the market environment, such as Total Supply, Total Demand, and Price, all of which are real values providing essential data on current market conditions. The Progress action, represented as an ordinal variable, informs the agent of market trends and how the competitors are going to react, indicating whether production levels are rising, stable, or declining, while Timestep\%100 tracks the current timestep within a cycle, enabling temporal awareness of the sinusoidal movements of supply and demand. Additionally, Competitors Quantities provides an array of production levels from competing agents seen in the previous timestep, allowing the agent to strategize based on competitor actions.

On the other hand, the action space is defined by the variable Units Produced, an integer variable where the agent selects the number of units to produce within the range $[0, 14]$, directly influencing supply levels, potential profit and other variables of the simulation. Finally, the reward function, represented by Producer profit, quantifies the performance of the agent based on the profit generated, which acts as feedback for optimizing future actions, we could also extract more complicated reward functions but we leave that for further work.

\begin{table}[h!]
\centering
\begin{tabular}{|l|l|l|p{6cm}|}
\hline
\textbf{Name} & \textbf{Role} & \textbf{Variable Type} & \textbf{Description} \\ \hline
Total Supply & Observation & Real & Total supply available in the market \\ \hline
Total Demand & Observation & Real & Total demand in the market at the current timestep \\ \hline
Progress action & Observation & Ordinal & Indicates market trend based on production (up, equal, or down) \\ \hline
Timestep\%100 & Observation & Integer & Current timestep within a 100-timestep cycle \\ \hline
Competitors Quantities & Observation & Real array & Production levels of each competitor \\ \hline
Price & Observation & Real & Current market price of the goods \\ \hline
Units Produced & Action & Integer & Number of units chosen for production by the agent \\ \hline
Producer profit & Reward & Real & The profit earned by the producer, used as the reward for the agent \\ \hline
\end{tabular}
\caption{Observation Space, Action Space, and Reward Function of the Agent.}
\label{table:agent_spaces}
\end{table}

In the following section, we will explain how we can train agents using the deep reinforcement learning methods described in previous section to estimate a successful policy in this complex market.

\section{Experiments and results}
We now detail the format of our experiments and interpret all the results of them. We begin by formalizing our research hypothesis, that is, the DRL agents are able to estimate a policy that is competitive in a complex market. More formally, we can validate this hypothesis through statistical hypothesis testing in the following way:

H0: $\mu_{DRL} \leq \mu_{random}$

H1: $\mu_{DRL} > \mu_{random}$

where $\mu_{DRL}$ is the mean performance of the policy estimated by the DRL agents and $\mu_{random}$ is the mean performance of a policy whose action is chosen at random at every timestep. Consequently, we would like to obtain empirical evidence that is not compatible with the null hypothesis, H0, that states that the mean performance of both methods is equal. Similarly, we want to validate an analogous hypothesis that can be formalized as:

H0: $\mu_{DRL} \leq \mu_{default}$

H1: $\mu_{DRL} > \mu_{default}$

where $\mu_{default}$ is the mean performance of a policy that always chooses to produce the same number of assets in every timestep independently on the market state represented in every observation. In both cases, we would like to reject the null hypothesis to accept the alternative hypothesis that jointly states that the performance of the DRL trained estimated policy performance is significantly superior to a random or fixed policy, and hence, it is effective in complex markets. We finally want to test that: 

H0: $\mu_{DRL} \leq 0$

H1: $\mu_{DRL} > 0$

that is, the agents obtain profits after a certain test period in the market. It is possible that the policy is better than a random or fixed policy but ending in red numbers, consequently, we also need to validate that the agents have profits at the end of the test simulation.

In order to obtain empirical evidence to support these claims and reject the null hypothesis, we design the following experiment using the environment described in the previous section. We implement a dummy vectorized environment of the market simulation where we trained $10$ agents using the PPO DRL algorithm from $[0,14]$ items that can be produced in each timestep. We also launch $15$ agents with a default policy that always produce an asset between $[0,14]$, each one corresponding to an asset number. In order to obtain evidence, we also initialize $5$ random agents that simply choose in every timestep their action at random. Finally, for completeness, we also test value based and combined value and policy based DRL with a DQN and a A2C agent. We train each DRL agents for $1.5$M timesteps using a learning rate of $10^{-4}$ and default hyperparameters. We represent the policy in the DRL agents with a multilayer perceptron neural network with the default StableBaselines3 configuration. For all the agents and configurations, we save different information that can be interpreted once the training period is finished. For evaluation purposes we launch all the agents $1000$ timesteps and record the cumulated profits that are the reward of every timestep to compare their performances. 

\begin{figure}[h!]
    \centering
    \includegraphics[width=\textwidth]{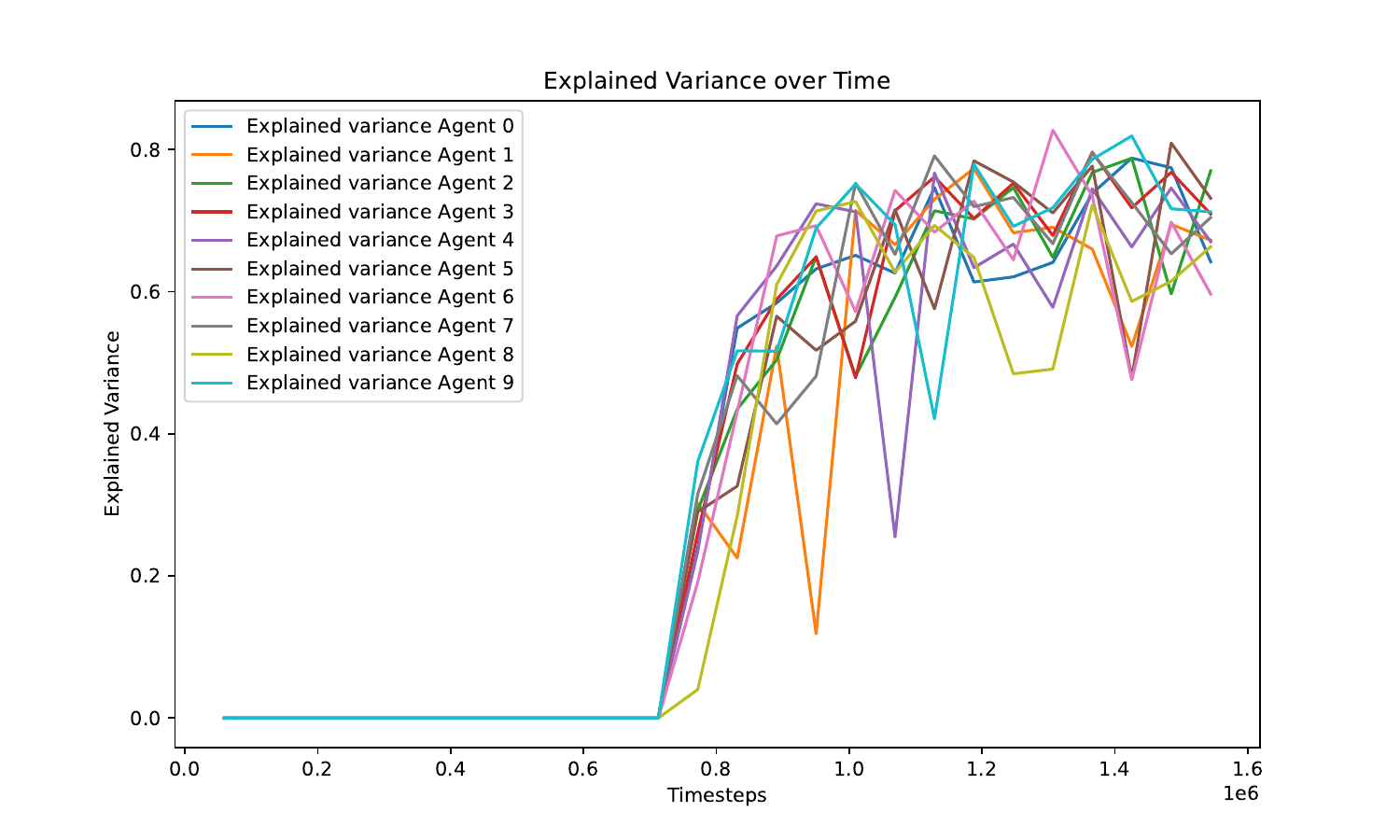}
    \caption{Explained Variance: This plot shows the explained variance over time, indicating how well the model captures the variance in the data.}
    \label{exp_var}
\end{figure}

First, we want to ensure that the DRL agents are effectively learning from the environment about the endogeneous variable of the experiment, the profits. We can see in Figure \ref{exp_var} the explained variance in the Y axis and the training timesteps in the X axis of every PPO agent. 

\begin{figure}[h!]
    \centering
    \includegraphics[width=\textwidth]{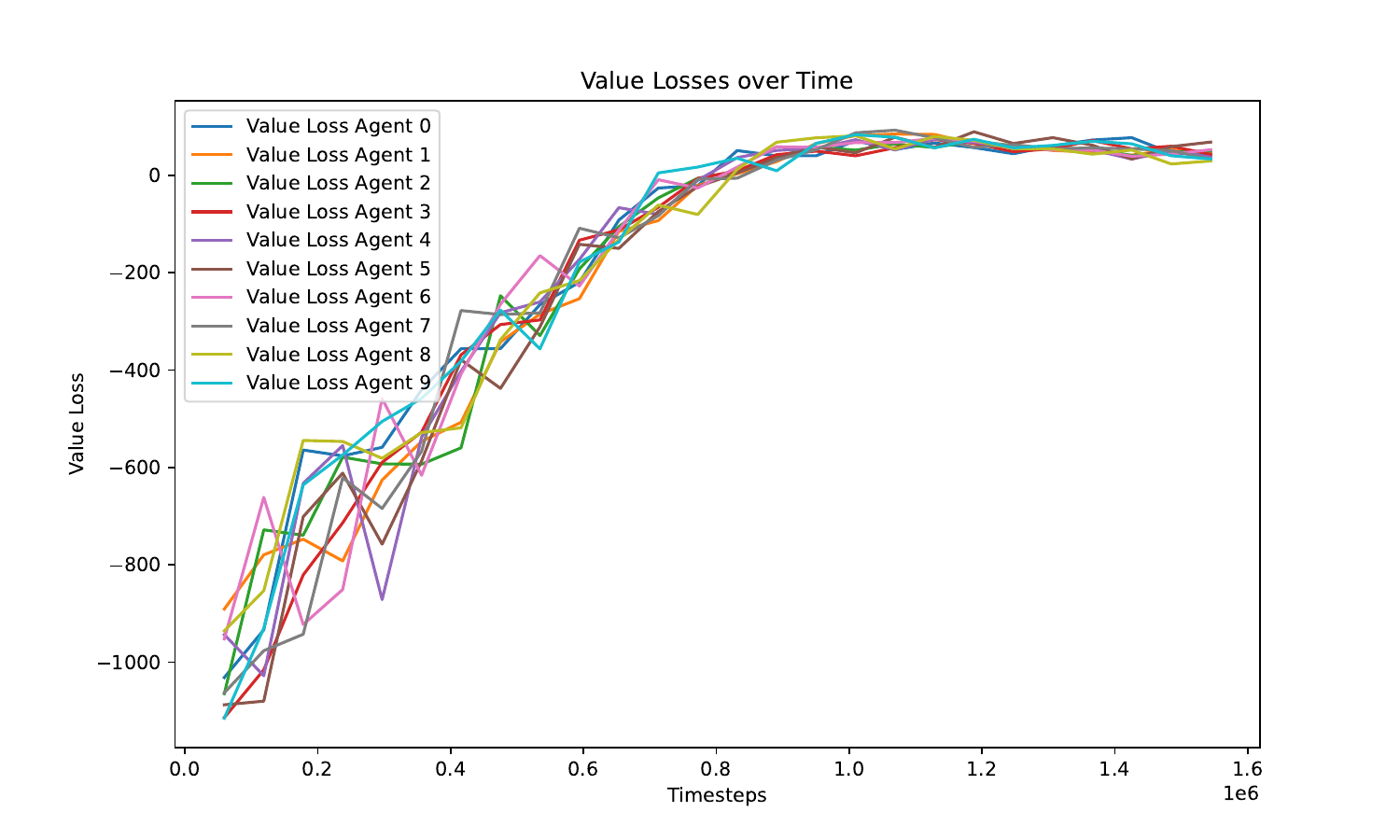}
    \caption{Value Loss: This plot shows the value loss over time, which helps monitor the training stability of the model.}
    \label{value_loss}
\end{figure}

We can see how the variability of the profits is starting to be explained from approximately $700$K training timesteps and how the agents are able to explain a $[60,80]$\% of the profits, which is an amazing result given that the action of the agent does not condition the profits alone and lots of variables that condition the profits are contaminated by noise. In prediction terms, we also see the same trend in Figure \ref{value_loss}, where we can see how the agents are able to better predict as the training process is done.

\begin{figure}[h!]
    \centering
    \includegraphics[width=\textwidth]{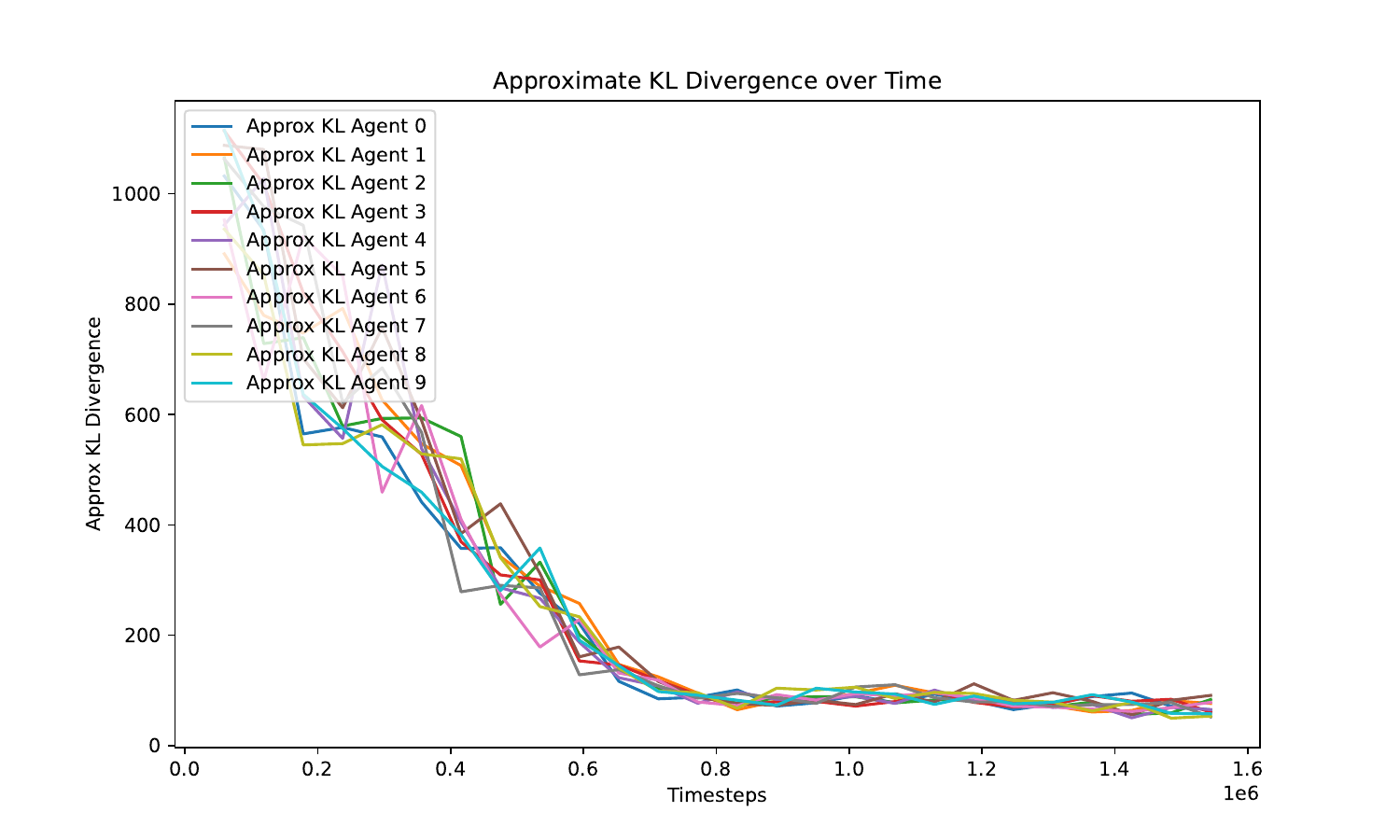}
    \caption{Approximate KL Divergence: This plot shows the approximate KL divergence values over time or iterations.}
    \label{kl_divergence}
\end{figure}

It is also interesting to check whether the changes in the policy distribution being learnt with respect to the previous policy in training time are lower as the training process goes on. This means that the training is more stable, converging into a stationary trajectory distribution. We can see in Figure \ref{kl_divergence} how this phenomenon happens as the KL divergence is lower as the training process continues until it is almost zero, meaning that changes in the policy distribution are not existent as the new states and actions performed to not add more information about the profits.

\begin{figure}[h!]
    \centering
    \includegraphics[width=\textwidth]{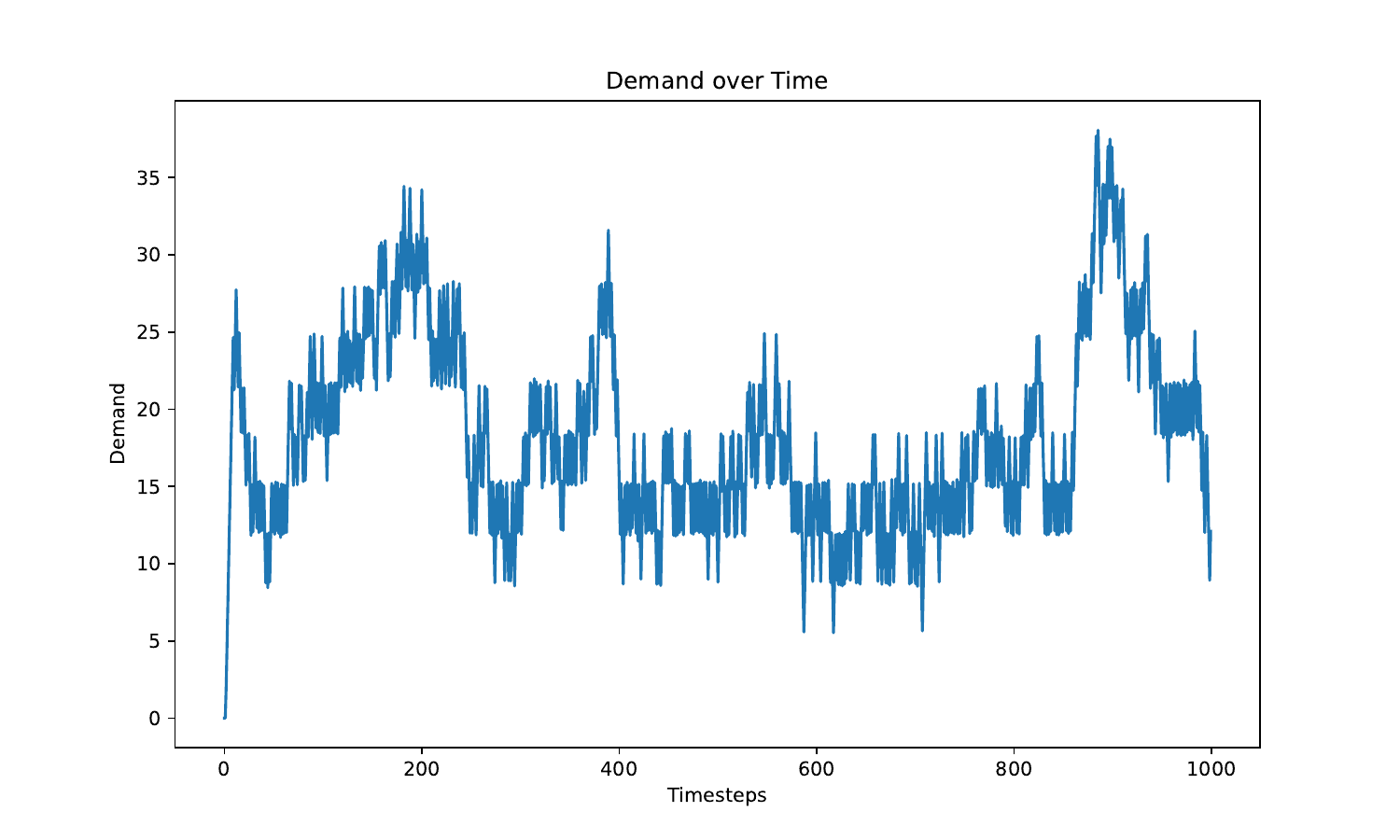}
    \caption{Demand over Timesteps: This plot shows the variation in demand across timesteps in the simulation.}
    \label{demand}
\end{figure}

Having analyzed the training process and successfully seen that the agents are learning and their policy is stable, we are now going to analyze the market dynamics. For example, it is interesting to analyze the demand in testing time, shown in Figure \ref{demand}. 

\begin{figure}[h!]
    \centering
    \includegraphics[width=\textwidth]{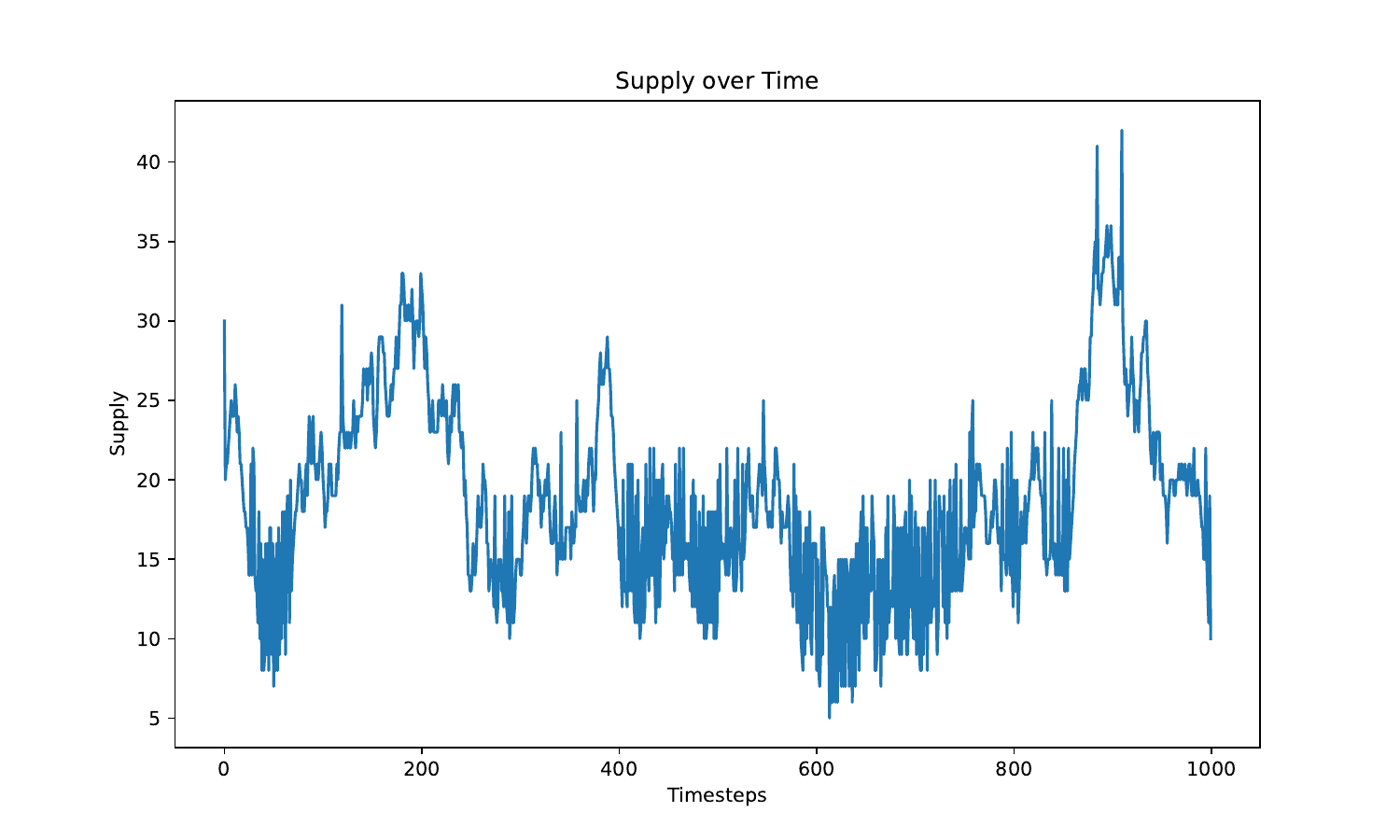}
    \caption{Supply over Timesteps: This plot shows the variation in supply across timesteps in the simulation.}
    \label{total_supply}
\end{figure}

We can see how at the beggining of the simulation the demand goes from 0 to a value of $25$ and how it oscillates as a function of the timesteps but it is also contaminated with noise, as the changes of the demand are not smooth, representing the complexities in the market. The same thing happens with the total supply for our agent shown in Figure \ref{total_supply}. 

\begin{figure}[h!]
    \centering
    \includegraphics[width=\textwidth]{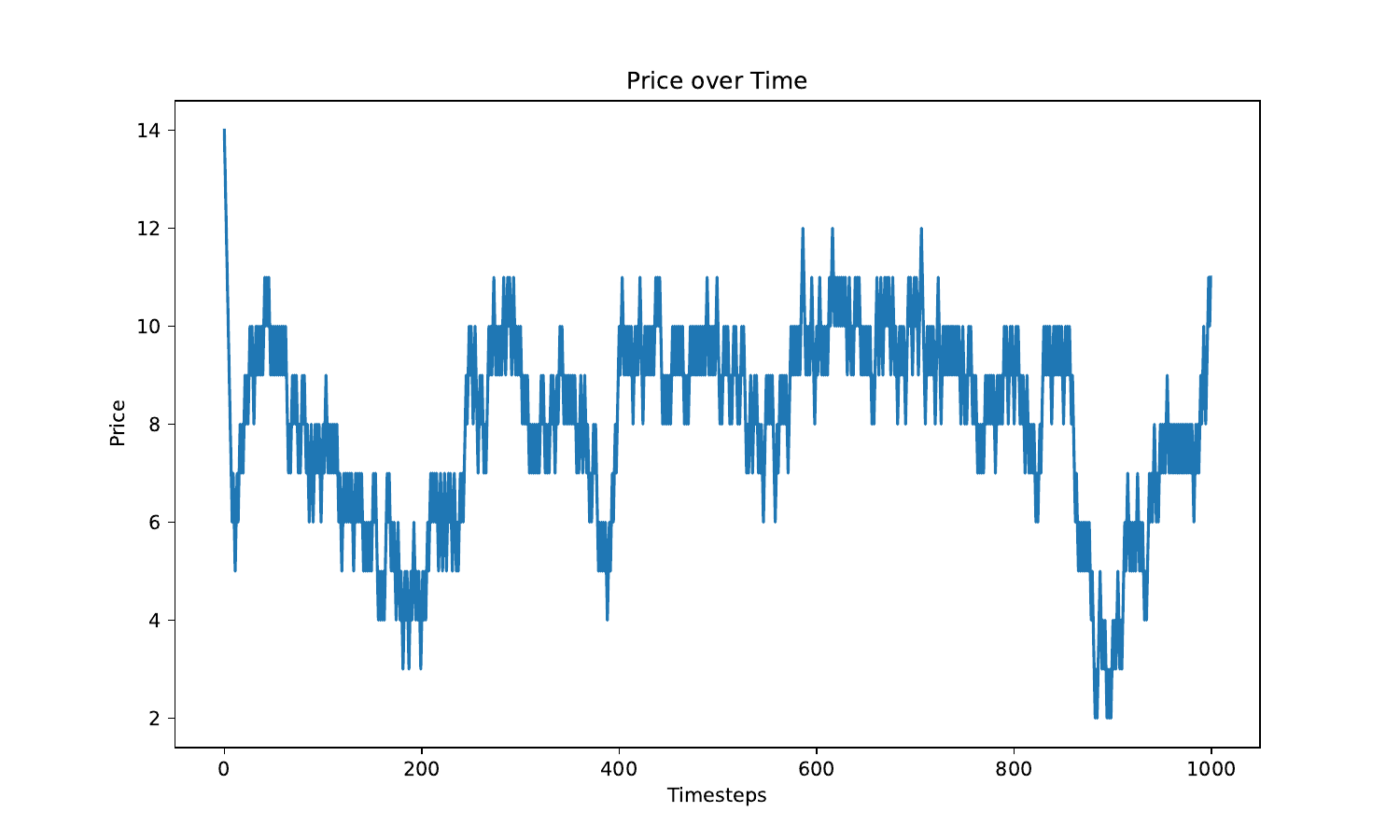}
    \caption{Price over Timesteps: This plot shows the variation in price across timesteps in the simulation.}
    \label{price}
\end{figure}

We can see that both figures are not necessarily equal, though variations are somewhat correlated, meaning that when the demand increases the supply of the provider of the agent also increases as it is aware that the demand of the agent product has grown, however it is not a direct correlation and it is also contaminated by noise although it has some stationary oscillations. Finally, we consider also interesting to study the price of the asset, shown in Figure \ref{price}.

We can see how the market dynamics influences the price and how it is also noisy, representing the complex behaviour of demand and supply. It also oscillates and it is somewhat negatively correlated with the supply, meaning that more amounts of the asset being available in the market makes the price of the asset lower, however it is not a strong correlation, representing complex dynamics in the market. 

Having studied the learning process of the DRL agents and the market dynamics, we are now able to illustrate the results of the profits accumulated by all the agents in test time to see whether we have obtained enough empirical evidence to reject the three null hypotheses associated with the research question of this manuscript, that is, that the agents are able to effectively learn a policy in a complex market. 

\begin{figure}[h!]
    \centering
    \includegraphics[width=\textwidth]{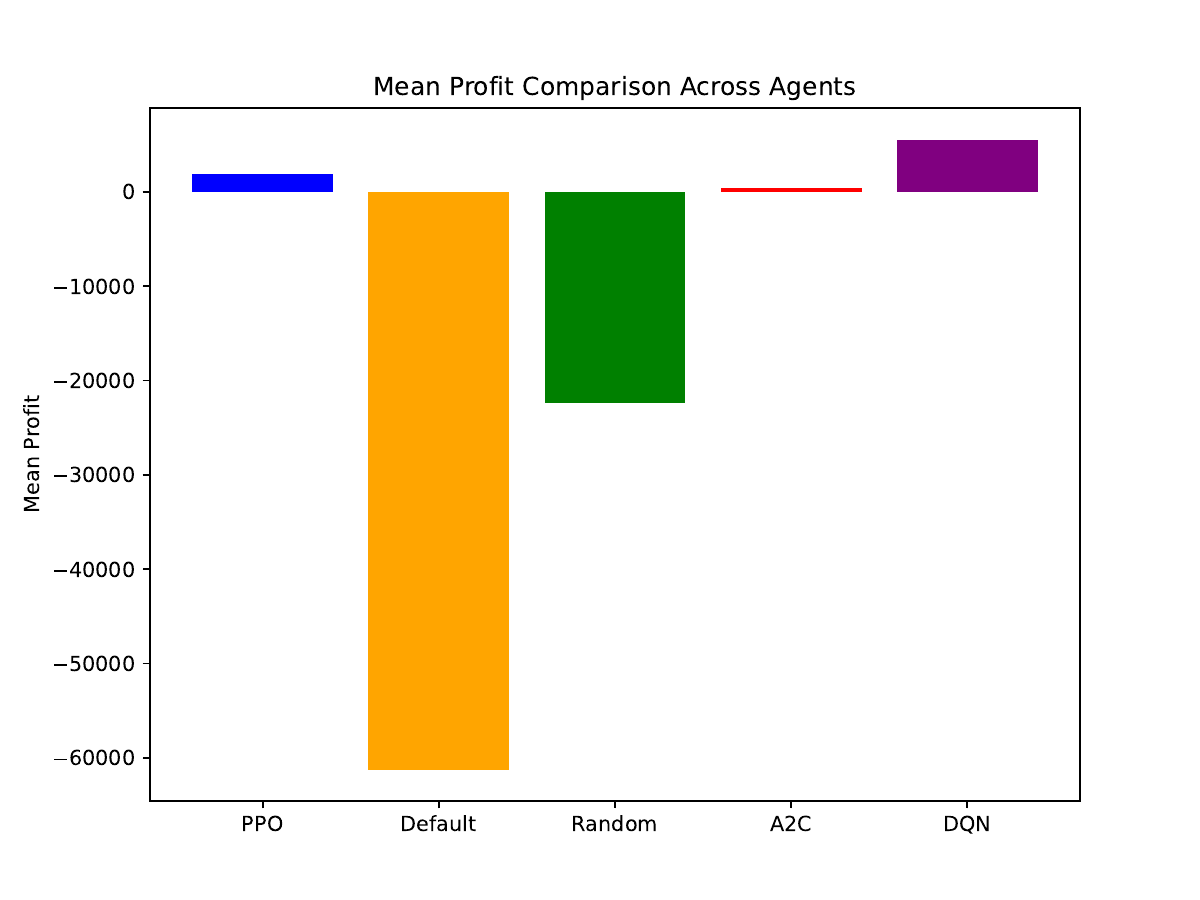}
    \caption{Mean Bar Plot: This bar plot shows the mean values of key metrics or variables.}
    \label{barplot}
\end{figure}

In Figure \ref{barplot} we show a mean profit comparison of all the agents trained in the experiment, the 10 PPO agents, the A2C agent, the DQN agent, the $15$ default agents and the random agents with exciting results.    

\begin{figure}[h!]
    \centering
    \includegraphics[width=\textwidth]{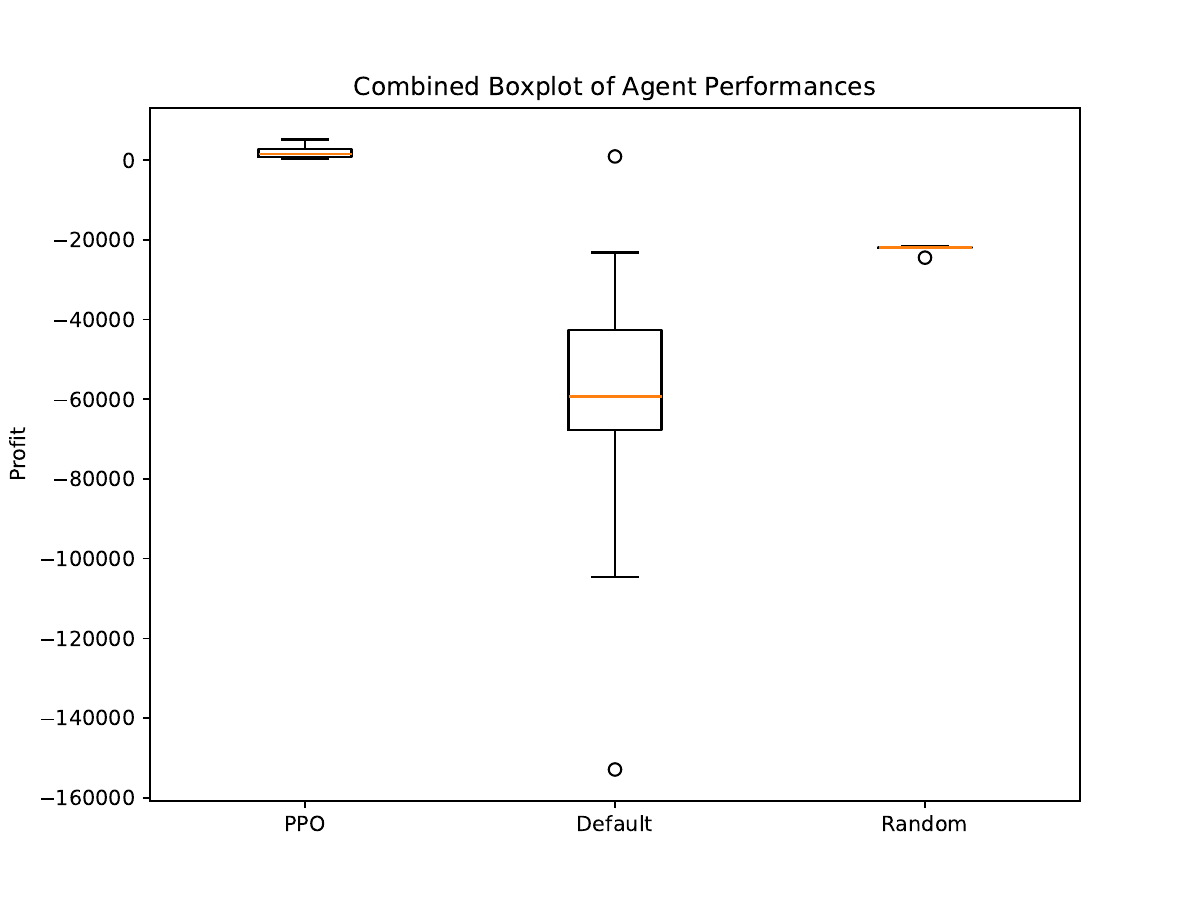}
    \caption{Combined Boxplot: This boxplot presents a combined summary of multiple data distributions or metrics.}
    \label{ppobox}
\end{figure}

Concretely, we can see how the DRL agents are the only agents that consistently obtain positive cumulative profits in all the experiments. Concretely, the minimum accumulated profits in the $1000$ timesteps of the experiment is $898.65$ but the maximum is $5129.20$. The PPO mean performance is $1942.89$ and the standard deviation is $1476$. As all the quantities fall in that range, we can safely assume that the DRL agents, in this training configuration, are going to have a mean positive profit, obtaining enough evidence to reject the null hypothesis that the profit is equal to zero and negative. It is interesting that the DQN agent has surprisingly obtained a good performance with respect to the PPO agent, our default chosen algorithm due to its popularity as we can see in the boxplot and lines of Figure \ref{ppobox}. 

\begin{figure}[h!]
    \centering
    \includegraphics[width=\textwidth]{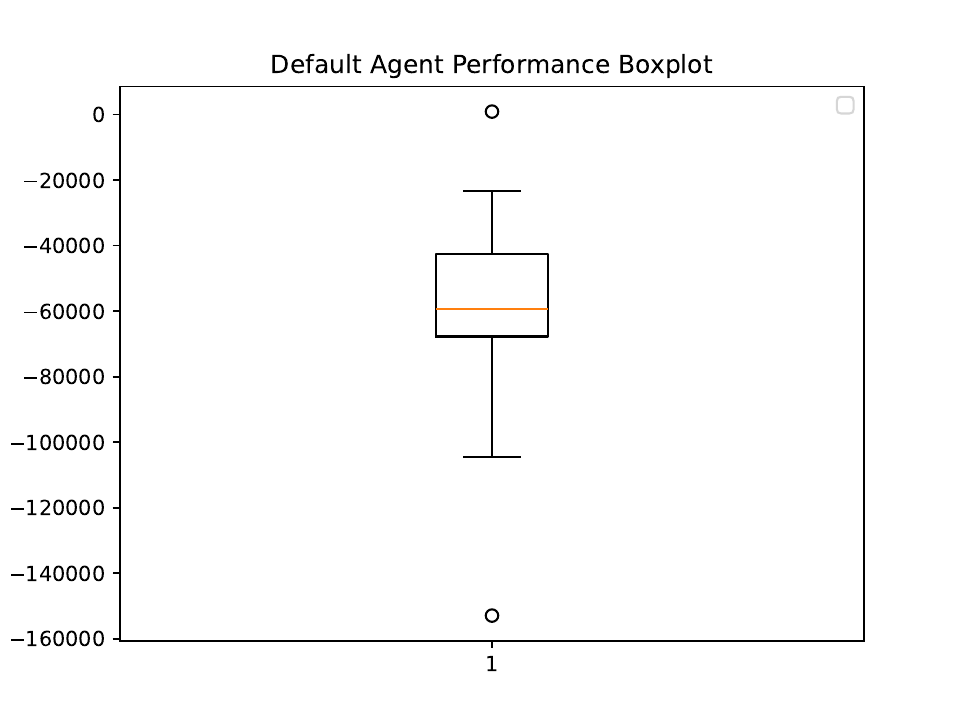}
    \caption{Default Boxplot: This boxplot illustrates the distribution of the default configuration or baseline.}
    \label{defagents}
\end{figure}

Having retrieved this evidence, we leave for further work, then, the hyperparameter tuning problem of the DRL algorithms and model selection as it is promising. We now focus on the first two hypothesis and provide evidence about them. We illustrate all the results of the random and default agents in Figure \ref{defagents}. 

\begin{figure}[h!]
    \centering
    \includegraphics[width=\textwidth]{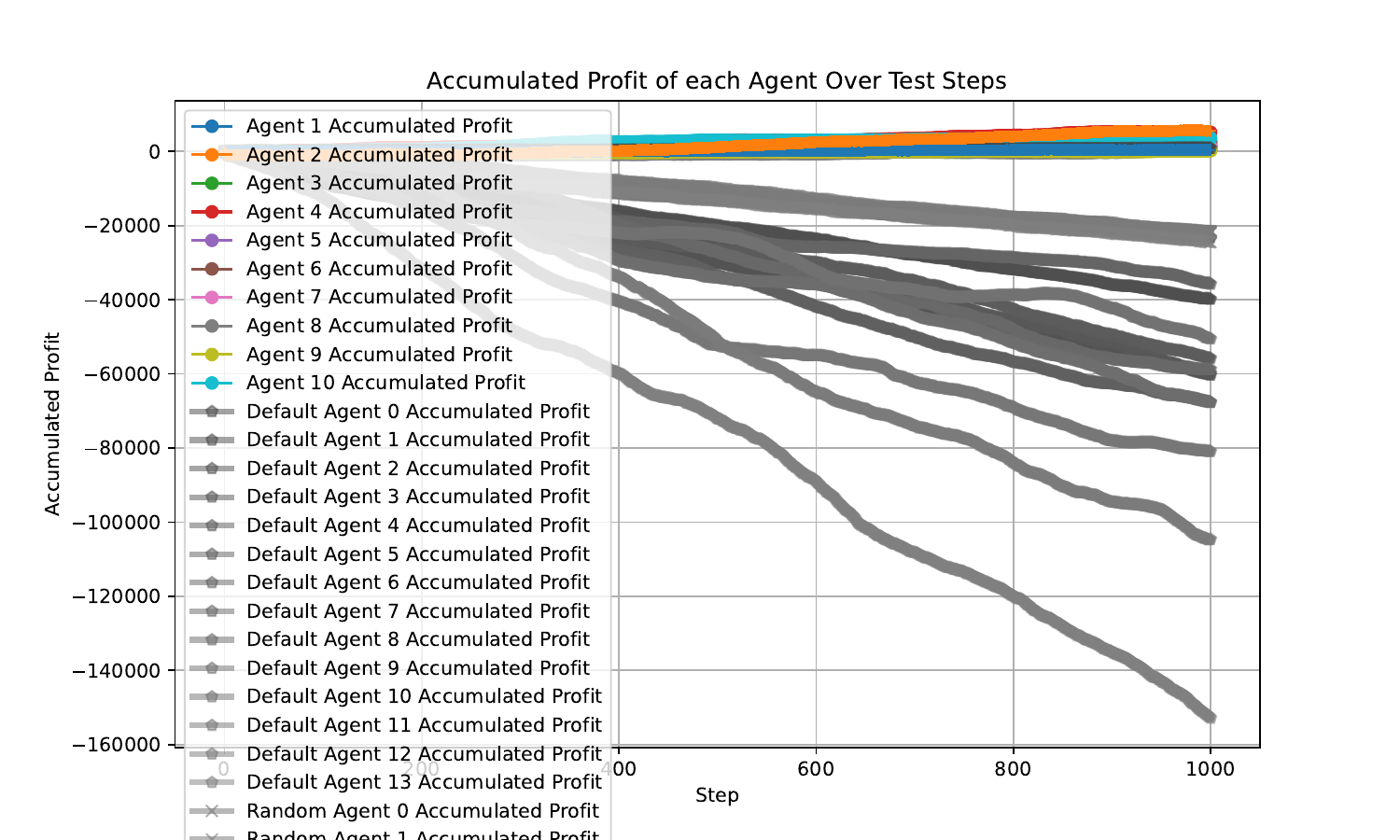}
    \caption{Profit Evolution: This plot depicts how the profit changes over time or iterations in the simulation.}
    \label{track}
\end{figure}

More concretely, the mean default performance is $-61217.42$ and the random performance is $-22333.67$, representing that our market simulation is hard and no easy policy can guarantee you a good result. Only the exception of consistently producing $1$ quantity of the asset generates a small $102.32$ profit after all the simulation, being consistently lower than the minimum of the DRL agents that is $898.65$. To see the evolution of the profit in testing time we also track the profits in Figure \ref{track}.

We can see how all the agents in gray that are the default and random agents are uniformly incurring in red numbers, some of then considerable losses, whereas the DRL agents are profitable. 

Finally to formalize the empirical evidence obtain we run a t-test that compares the performance of DRL with respect to random obtaining a t-statistic of $33.13$ and an associated p-value of almost zero. Analogously, we obtained a t-statistic of $6.47$ in the case of default comparison also with an associated p-value lower than $10^{-5}$. Consequently, with a significance level $\alpha=0.001$ we reject the null hypotheses and accept the alternative hypotheses that establish that DRL agents performance is better than a naive policy consisting on random or default actions. Moreover we have shown how the variability of the profits is very well explained by the agents, how the training process is stable according to the KL divergence and has good performance behaviour according to the value loss and the cumulated profits. Given all this accumulated empirical evidence, we consider satisfied our research question whose claim is that DRL agents successfully estimate competitive policies in a complex market.   

\section{Conclusions and Further Work}
This paper demonstrates how DRL agents can be effectively trained to estimate optimal production policies for a specific item, enabling competitiveness in a highly complex market environment. In particular, our approach is the first DRL attempt that illustrates the way for more sophisticated actions within the agent's strategy, accommodating various complementary and substitute products that influence market dynamics. By broadening the scope of available actions or simulating a multi agent system, the agents will gain flexibility to adapt its production policy in response to the evolving landscape of interconnected goods. 

We have seen that we have used point estimations for the hyperparameters of the DRL agents, whose probability of being the optimum in a continous hyperparameter space is 0. Consequently, we believe that this approach will significantly benefit for performing DRL hyperparameter tuning using Bayesian optimization (BO) \cite{frazier2018bayesian, afshar2022automated}, as only a few BO evaluations may be reasonable to do due to computational resources.

Finally, in order to satisfy the investors, an essential aspect of our methodology is making the resulting policies explainable. Using recent advancements in DRL explainable AI \cite{de2024explainable}, we can interpret the agent's decision-making processes, allowing stakeholders to understand the rationale behind each production adjustment. This interpretability is crucial for building trust and transparency, especially in complex systems where black-box decisions can be challenging to justify.

\bibliographystyle{unsrt}
\bibliography{bibliography}

\end{document}